\pdfoutput=1

\documentclass[11pt]{article}

\usepackage{acl}

\usepackage{times}
\usepackage{latexsym}
\usepackage{graphicx}
\usepackage{amsmath}
\usepackage{multirow}
\usepackage{amsthm}
\usepackage{amsfonts}
\usepackage{bm}
\usepackage{booktabs}
\usepackage{algorithm}
\usepackage{algorithmic}
\usepackage{verbatim}

\usepackage[T1]{fontenc}

\usepackage[utf8]{inputenc}

\usepackage{microtype}

\title{Complex Evolutional Pattern Learning for Temporal Knowledge Graph Reasoning}

\author{Zixuan Li\textsuperscript{1,2,3}\thanks{This work was done while the first author was doing internship at Baidu Inc.}, Saiping Guan\textsuperscript{1,2},
Xiaolong Jin\textsuperscript{1,2}, Weihua Peng\textsuperscript{3}, Yajuan
Lyu\textsuperscript{3},\\
\textbf{ Yong Zhu\textsuperscript{3}, Long Bai\textsuperscript{1,2}, Wei Li\textsuperscript{3}, Jiafeng
Guo\textsuperscript{1,2}, Xueqi Cheng\textsuperscript{1,2}} \\
 \textsuperscript{1}School of Computer Science and Technology, University of Chinese Academy of Sciences; \\
 \textsuperscript{2}CAS Key Laboratory of Network Data Science and Technology, \\Institute of Computing Technology, Chinese Academy of Sciences;
 \textsuperscript{3}Baidu Inc. \\
 \texttt{\{lizixuan,guansaiping,jinxiaolong\}@ict.ac.cn}\\
 \texttt{\{pengweihua,lvyajuan,zhuyong\}@baidu.com}
 }

\begin{document}
\maketitle
\begin{abstract}
  A Temporal Knowledge Graph (TKG) is a sequence of KGs corresponding to
  different timestamps. TKG reasoning aims to predict potential facts in the
  future given the historical KG sequences. One key of this task is to mine and
  understand evolutional patterns of facts from these sequences. The
  evolutional patterns are complex in two aspects, length-diversity and
  time-variability. Existing models for TKG reasoning focus on modeling fact
  sequences of a fixed length, which cannot discover complex evolutional
  patterns that vary in length. Furthermore, these models are all trained
  offline, which cannot well adapt to the changes of evolutional patterns from
  then on. Thus, we propose a new model, called Complex Evolutional Network
  (CEN), which uses a length-aware Convolutional Neural Network (CNN) to handle
  evolutional patterns of different lengths via an easy-to-difficult curriculum
  learning strategy. Besides, we propose to learn the model under the online
  setting so that it can adapt to the changes of evolutional patterns over time.
  Extensive experiments demonstrate that CEN obtains substantial performance
  improvement under both the traditional offline and the proposed online
  settings.
\end{abstract}

\section{Introduction}

Temporal Knowledge Graph
(TKG)~\cite{boschee2015icews,gottschalk2018eventkg,gottschalk2019eventkg,
zhao2020event} has emerged as a very active research area over the last few
years. Each fact in TKGs is a quadruple \emph{(subject, relation, object,
timestamp)}. A TKG can be denoted as a sequence of KGs with timestamps, each of
which contains all facts at the corresponding timestamp. TKG reasoning aims to
answer queries about future facts, such as \emph{(COVID-19, New medical case
occur, ?, 2022-1-9)}. 

To predict future facts, one challenge is to dive deep into the related
historical facts, which reflect the preferences of the related entities and
affect their future behaviors to a certain degree. Such facts, usually
temporally adjacent, may carry informative sequential patterns, called
evolutional patterns in this paper. For example, [\emph{(COVID-19, Infect, A,
2021-12-21), (A, Discuss with, B, 2021-12-25), (B, Go to, Shop, 2021-12-28)}] is
an informative evolutional pattern for the above query implied in historical
KGs. There are two kinds of models to model evolutional patterns, namely,
query-specific and entire graph based models. The first kind of
models~\cite{jin2020recurrent,li2021search,
sun2021timetraveler, han2020explainable, han2021temporal, zhu2021learning} extract
useful structures (i.e., paths or subgraphs) for each individual query from the
historical KG sequence and further predict the future facts by mining
evolutional patterns from these structures. This kind of models may
inevitably neglect some useful evolutional patterns. Therefore, the entire graph
based models~\cite{deng2020dynamic, li2021search} take a sequence of entire
KGs as the input and encode evolutional patterns among them, which exhibit
superiority to the query-specific models. 

However, they all ignore the length-diversity and time-variability of
evolutional patterns. \textbf{Length-diversity}: The lengths of evolutional
patterns are diverse. For example, [\emph{(COVID-19, Infect, A, 2021-12-21), (A,
Discuss with, B, 2021-12-25), (B, Go to, Shop, 2021-12-28)}] is a useful
evolutional pattern of length 3 to predict the query \emph{(COVID-19, New
medical case occur, ?, 2022-1-9)} and [\emph{(COVID-19, Infect, A, 2021-12-21),
(A, Go to, Shop, 2021-12-30)}] is also a useful evolutional pattern of length 2
for this query. Previous models extract evolutional patterns of a fixed length,
which cannot handle evolutional patterns of diverse lengths.
\textbf{Time-variability}: Evolutional patterns change over time. For example,
\emph{(COVID-19, Infect, A, 2019-12-9)} and \emph{(COVID-19, Infect, A,
2022-1-9)} may lead to different results due to the wide usage of the COVID-19
vaccines. Previous models learn from the historical training data, which fail in
modeling the time-variability of evolutional patterns after that.

Upon the above observations, we propose Complex Evolutional Network (CEN) to
deal with the above two challenges. For length-diversity, CEN learns evolutional
patterns from historical KG sequences of different lengths via an Relational
Graph Neural Network (RGCN) based KG sequence encoder and a length-aware
Convolutional Neural Network (CNN) based evolutional representation decoder.
Besides, the model is trained via an easy-to-difficult curriculum learning
strategy incrementally according to the length of KG sequences. For
time-variability, we learn CEN under an online setting and combine CEN with a
temporal regularization unit to alleviate the catastrophic forgetting
problem~\cite{1989Catastrophic}.

In general, this paper makes the following contributions:
 \begin{itemize}
  \item We address, for the first time, the problems of length-diversity and
  time-variability of evolutional patterns for TKG reasoning.

  \item For length-diversity, we propose a length-aware CNN to learn evolutional
  patterns with different lengths in a curriculum learning manner. For
  time-variability, we propose to learn the model under an online setting to
  adapt to the changes of evolutional patterns.
  
  \item Experiments demonstrate that the proposed CEN model achieves better
  performance on TKG reasoning under both the traditional offline and the
  proposed online settings.
\end{itemize}

\section{Related Work}
The TKG reasoning task primarily has two settings, interpolation and
extrapolation. This paper focus on the extrapolation setting. In what
follows, we will introduce related work on both settings: 

\textbf{TKG Reasoning under the interpolation setting.}
This setting aims to complete the missing facts at past timestamps
\cite{jiang2016encoding,leblay2018deriving, dasgupta2018hyte,garcia2018learning,
goel2020diachronic, wu2020temp}. For example, TTransE \cite{leblay2018deriving}
extends TransE \cite{bordes2013translating} by adding the temporal constraints;
HyTE \cite{dasgupta2018hyte} projects the entities and relations to time-aware
hyperplanes to generate representations for different timestamps. Above all,
they cannot obtain the representations of the unseen timestamps and are not
suitable for the extrapolation setting.

\textbf{TKG Reasoning under the extrapolation setting} This setting aims to
predict facts at future timestamps, which can be categorized into two groups:
query-specific and entire graph based models. Query-specific models
focus on modeling the query-specific history. For example,
RE-NET~\cite{jin2020recurrent} captures the evolutional patterns implied in the
subgraph sequences of a fixed length specific to the query.
CyGNet~\cite{zhu2021learning} captures repetitive patterns by modeling
repetitive facts. xERTE~\cite{han2020explainable} learns to find the
query-related subgraphs of a fixed hop number. CluSTeR~\cite{li2021search} and
TITer ~\cite{sun2021timetraveler} both adopt reinforcement learning to discover
evolutional patterns in query-related paths of a fixed length. Unlike the
query-specific models, entire graph based models encode the latest
historical KG sequence of a fixed-length. RE-GCN~\cite{li2021temporal} captures
the evolutional patterns into the representations of all the entities by
modeling KG sequence of a fixed-length at lastest a few timestamps.
Glean~\cite{deng2020dynamic} introduces event descriptions to enrich the
information of the entities.
\vspace{-2mm}

\section{Problem Formulation}\label{Formulation} A TKG $G =\{G_{1}, G_{2}, ...,
G_{t}, ...\}$, where $G_{t}=(\mathcal{V}, \mathcal{R}, \mathcal{E}_{t})$, is a
directed multi-relational graph. $\mathcal{V}$ is the set of entities,
$\mathcal{R}$ is the set of relations, and $\mathcal{E}_{t}$ is the set of facts
at timestamp $t$. The TKG reasoning task aims to answer queries like $(s, r,?,
t_q)$ or $(?, r, o, t_q)$ with the historical KG sequence $\{G_{1}, G_{2},
...,G_{t_q-1}\}$ given, where $s,o\in\mathcal{V}$, $r\in\mathcal{R}$ and $t_q$
are the subject/object entity, the relation and the query timestamp,
respectively. Following ~\citet{jin2020recurrent}, KGs from timestamps $1$ to
$T_1$, $T_1$ to $T_2$, $T_2$ to $T_3$ ($T_1<T_2<T_3$) are used as the training,
validation and test sets, respectively. Under the traditional offline setting,
models are trained only using the training set ($t_q\le T_1$), while under the
online setting, the model will be updated by KGs before $t_q$ ($T_1<t_q\le T_3$)
continually. Without loss of generality, we describe our model as predicting the
missing object entity.

\section{Methodology}
We propose CEN to deal with the length-diversity and time-variability challenges
of evolutional pattern learning for TKG reasoning. Specifically, CEN consists of
a basic model as well as a curriculum learning strategy for the former challenge
and an online learning strategy for the latter challenge.


\begin{figure}[tbp]  
  \centering
  \includegraphics[width=3in]{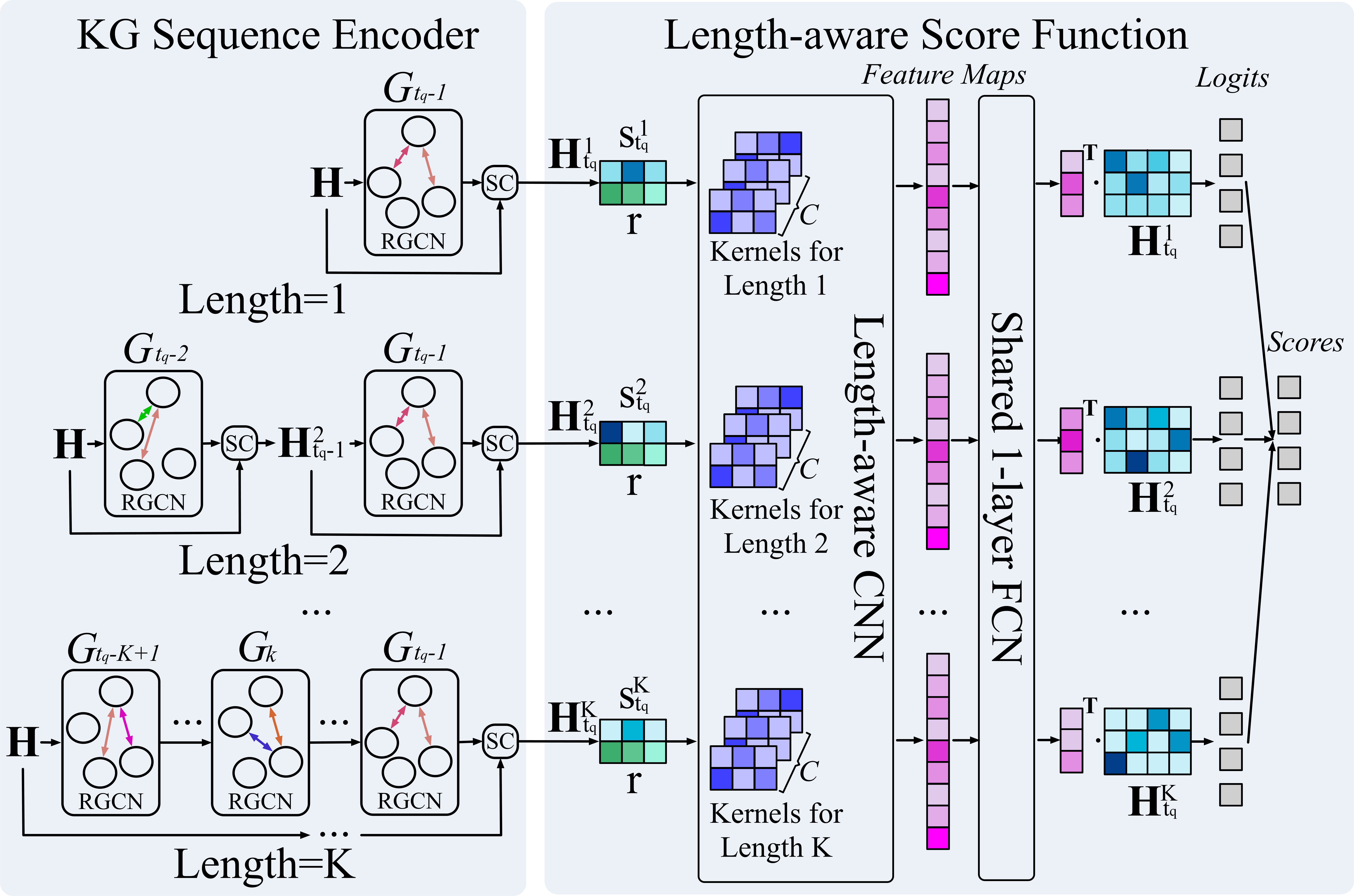}
  \caption{An diagram of the basic CEN model.}
  \vspace{-4mm}
  \label{fig:framework}
  \end{figure}

\subsection{\bf Basic CEN Model} 
As shown in Figure~\ref{fig:framework}, the basic model of CEN contains a KG
sequence encoder and an evolutional representation decoder. The KG sequence
encoder encodes the latest historical KG sequences of different lengths to
corresponding evolutional representations of entities. Then, the
evolutional representation decoder calculates the scores of all entities for the
query based on these representations.

{\bf KG Sequence Encoder.} Its inputs include the lastest historical KG
sequences of lengths from $1$ to $K$, initial representations of entities
$\mathbf{H}\in\mathbb{R}^{|\mathcal{V}|\times d}$ and relation representations
$\mathbf{R}\in \mathbb{R}^{|\mathcal{R}|\times d}$, where $d$ is the dimension
of the representations. Take the KG sequence of length $k=2$ for example, for
each KG in the input sequence $\{G_{t_q-2}, G_{t_q-1}\}$, it iteratively
calculates the evolutional representations of entities $\mathbf{H}^{2}_{t}$ at
the corresponding timestamps $t\in\{t_q-1, t_q\}$ as follows:
\begin{align}
\hat{\mathbf{H}}^{2}_{t}&=RGCN( \mathbf{H}^{2}_{t-1}, \mathbf{R}, G_{t-1}),\\
\mathbf{H}^{2}_{t} &=SC(\hat{\mathbf{H}}^{2}_{t}, \mathbf{H}^{2}_{t-1}),
\end{align}
where $RGCN(\cdot)$ and $SC$ denote the shared RGCN layer and the skip
connection unit proposed in RE-GCN~\cite{li2021temporal}. For the initial
timestamp $t_q-1$, $\mathbf{H}^{2}_{t_q-2}$ is set to $\mathbf{H}$. $\mathbf{R}$
is shared across timestamps, which is different from RE-GCN. By reusing the
encoder for KG sequences of different lengths, we obtain $K$ entity evolution
representations at the query timestamp: $\{\mathbf{H}_{t_q}^{1}, ..., \mathbf{H}_{t_q}^{k}, ...,
\mathbf{H}_{t_q}^K\}$.


\begin{figure}[tbp]  
  \centering
  \includegraphics[width=2.9in]{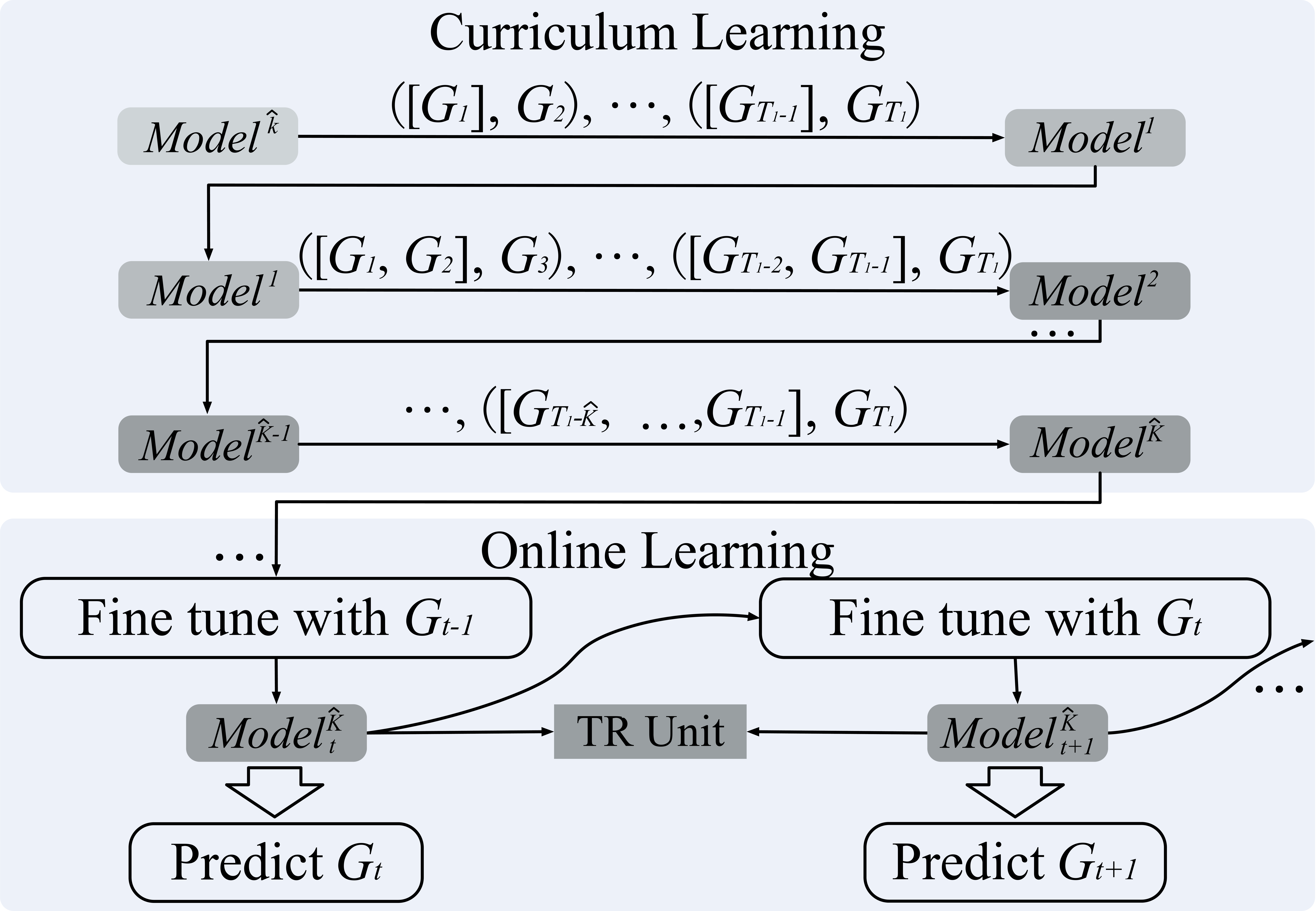}
  \caption{The learning procedure of the proposed model.}
  \vspace{-4mm}
  \label{fig:online}
  \end{figure}

{\bf Evolutional Representation Decoder.} Multiple evolutional representations
contain evolutional patterns of multiple lengths. To distinguish the influences
of the length-diverse evolutional patterns, we design a length-aware CNN, which
uses $K$ separate channels to model the above $K$ evolutional representations. 
Specifically, for a query $(s, r, ?, t_q)$, the representations of $s$
($\mathbf{s}^{1}_{t_q}, ..., \mathbf{s}^{k}_{t_q}, ..., \mathbf{s}^{K}_{t_q}$)
and $r$ ($\mathbf{r}$) are looked up from multiple representations of entities
$\{\mathbf{H}_{t_q}^{1}, ..., \mathbf{H}_{t_q}^{k}, ..., \mathbf{H}_{t_q}^K\}$
and the shared relation representations $\mathbf{R}$. For historical KG sequence
of length $k$, $k^{th}$ channel with $C$ different kernels of size $2\times M$
is used to decode the concatenation of $\mathbf{s}^{k}_{t_q}$ and $\mathbf{r}$.
Specifically, the feature maps are calculated as below, 
\begin{align}
\vspace{-3mm}
\mathbf{m}_c^{k}(s, r, t_q) = Conv_{2D}(\mathbf{w}_c^{k}, [\mathbf{s}^{k}_{t_q};\mathbf{r}]),
\vspace{-3mm}
\end{align}
where $Conv_{2D}$ denotes the 2D convolution operation, $\mathbf{w}_c^{k}$
($0\le c<C$) are the trainable parameters in $c^{th}$ kernel of $k^{th}$ channel
and $\mathbf{m}_c^{k}(s, r, t_q)\in\mathbb{R}^{1\times d}$. 
After that, it concatenates the output vectors from $C$ kernels yielding a
vector: $\mathbf{m}^k(s, r, t_q) \in \mathbb{R}^{C\times d}$. For $K$ channels,
it outputs a list of vectors: [$\mathbf{m}^1(s, r, t_q)$,... , $\mathbf{m}^k(s,
r, t_q)$,..., $\mathbf{m}^K(s, r, t_q)$]. Then, each vector is fed into a shared
1-layer Fully Connected Network (FCN) with $\mathbf{W}_3\in \mathbb{R}^{Cd\times
d}$ as its parameters and the final score of a candidate entity $o$ is the sum
of the logits from multiple evoltional representations:
$\sum_{k=1}^{K}\mathbf{m}^{k}(s, r, t_q)\mathbf{W}_3\mathbf{o}^{k}$, where
$\mathbf{o}^k$ is the evolutional representation of length $k$ for $o$. Then we
seen it as a multi-class learning problem and use the cross-entropy as its
objective function.

\subsection{Curriculum Learning for Length-diversity}
Longer historical KG sequences contain more historical facts and longer
evolutional patterns, which is more challenging to learn. Similar to human
learning procedures, the models can benefit from an easy-to-difficult
curriculum. Besides, how to choose the maximum length of evolutional patterns is
vital to CEN. Thus, we design the curriculum learning strategy to learn the
length-diverse evolutional patterns from short to long and adaptively select the
optimal maximum length $\hat{K}$. As shown at the top of
Figure~\ref{fig:online}, we start from the minimum length $\hat{k}$ ($\hat{k}=1$
for example) and gradually move on to longer history in the training set. The
model stops the curriculum and gets the optimal $\hat{K}$ when the MRR metric
decreases or the length is up to maximum length $K$. Note that, curriculum
learning is conducted under the traditional offline setting and
$Model^{\hat{K}}$ is used as the pre-trained model for online learning.

\subsection{Online Learning for Time-variability}

To handle the time-variability of evolutional patterns, one simple and direct
method is to update the model according to the newly occurred facts. Thus, as
shown in the bottom of Figure~\ref{fig:online}, for timestamp $t+1$ ($T_1 < t+1
< T_3$), $Model^{\hat{K}}_{t}$ is fine-tuned to get $Model^{\hat{K}}_{t+1}$ by
predicting the facts in the KG at the last timestamp $G_{t}$ with historical
KG sequences as inputs. Furthermore, to balance the knowledge of new evolutional
patterns and the existing ones, we use a Temporal Regularization unit (TR
unit)~\cite{daruna2021continual, 2021TIE}. We apply an $L2$ regularization
constraint between two temporally adjacent models to smooth the drastic change
of the parameters.

\subsection{Analysis on Computational Complexity}
We analyze the computational complexity of CEN. We view the
computational complexities of the RGCN unit and ConvTransE as constants. Then,
the time complexity of the RGCN at a timestamp $t$ is ${O}(|\mathcal{E}|)$,
where $|\mathcal{E}|$ is the maximum number of facts at timestamps in history.
As we unroll $m$ $(m=\hat{K}-\hat{k})$ sequences, the time complexity of the KG
sequence encoder is finally ${O}({m}^2|\mathcal{E}|)$. Thus, the time complexity
of CEN is ${O}({m}^2|\mathcal{E}|+m)$.

\begin{table}
  \small
  \centering
  \begin{tabular}{lrrrrrr}
  \toprule
  Datasets           &ICEWS14   &ICEWS18  &WIKI  \\
  \midrule
  $\#\mathcal{E}$              &6,869    &23,033  &12,554   \\
  $\#\mathcal{R}$               &230     &256   &24     \\
  $\#{Train}$     &74,845   &373,018  &539,286  \\
  $\#{Valid}$      &8,514    &45,995  &67,538   \\
  $\#{Test}$      &7,371    &49,545   &63,110 \\
  $T_3$        &1 day   &1 day   &1 year\\
  
  \bottomrule
  \end{tabular}
  \vspace{-2mm}
  \caption{Statistics of the datasets. $\#{Train}$, $\#{Valid}$, $\#{Test}$ are the
  numbers of facts in the training, validation and test sets. }
  \vspace{-4mm}
  \label{table:datasets}
  \end{table}

\begin{table*}[htb]
  \small
  \centering
  \begin{tabular}{lcccccccccccc}

  \toprule
  \multirow{2}{*}{Model} &\multicolumn{4}{c}{ICEWS14} &\multicolumn{4}{c}{ICEWS18} &\multicolumn{4}{c}{WIKI}\\
  \cmidrule(r){2-5}  \cmidrule(r){6-9} \cmidrule(r){10-13}  &MRR &H@1 &H@3 &H@10 &MRR &H@1 &H@3 &H@10 &MRR &H@1 &H@3 &H@10\\
  \midrule
  {CyGNet}\!\!\!       &35.05  &25.73  &39.01  &53.55     &24.93  &15.90  &28.28 &42.61  &33.89  &29.06  &36.10 &41.86   \\
  {RE-NET}\!\!\!       &36.93  &26.83  &39.51  &54.78     &28.81  &19.05  &32.44 &47.51  &49.66  &46.88  &51.19 &53.48   \\
  {xERTE}\!\!\!        &40.02  &32.06  &44.63  &56.17     &29.31  &21.03  &33.51 &46.48  &71.14  &68.05  &76.11 &79.01  \\
  {TG-Tucker}\!\!\!    &-     &-     &-     &-    &28.68  &19.35  &32.17 &47.04  &50.43  &48.52  &51.47 &53.58    \\
  {TG-DistMult}\!\!\!  &-     &-     &-     &-     &26.75  &17.92  &30.08 &44.09  &51.15  &49.66  &52.16 &53.35   \\
  {TITer}\!\!\!        &40.97 &\textbf{32.28} &45.45 &57.10  &29.98  &\textbf{22.05}  &33.46 &44.83  &75.50  &72.96  &77.49 &79.02     \\
  {RE-GCN}\!\!\!        &40.39 &30.66 &44.96 &59.21  &30.58  &21.01  &34.34 &48.75  &77.55  &73.75  &80.38 &83.68     \\
  {CEN}\!\!\!    &\textbf{42.20}   &32.08 &\textbf{47.46} &\textbf{61.31}  
  &\textbf{31.50} &21.70 &\textbf{35.44}  &\textbf{50.59}
  &\textbf{78.93} &\textbf{75.05} &\textbf{81.90} &\textbf{84.90}   \\
  \bottomrule
  \end{tabular}
  \vspace{-2mm}
  \caption{Experimental results on TKG reasoning (in percentage) under the offline setting.}
  \vspace{-2mm}
  \label{table:offline_results}
  \end{table*}

  \begin{table*}[htb]
  \small
  \centering
  \begin{tabular}{lcccccccccccc}
  \toprule
  \multirow{2}{*}{Model} &\multicolumn{4}{c}{ICEWS14} &\multicolumn{4}{c}{ICEWS18} &\multicolumn{4}{c}{WIKI} \\
  \cmidrule(r){2-5}  \cmidrule(r){6-9} \cmidrule(r){10-13}
  &MRR &H@1 &H@3 &H@10 &MRR &H@1 &H@3 &H@10 &MRR &H@1 &H@3 &H@10\\
  \midrule
  {CEN(-TR)}\!\!\!    &39.28   &30.05  &43.58 &57.01  &31.11 &21.41 &35.09 &50.27   & \textbf{81.92}  &\textbf{77.93} &\textbf{85.23} &\textbf{87.63}       \\
  {CEN}\!\!\!         &\textbf{43.34}   &\textbf{33.18} &\textbf{48.49} &\textbf{62.58}  &\textbf{32.66} &\textbf{22.55} &\textbf{36.81} &\textbf{52.50}     & 79.67  &75.63 &83.00 &85.58    \\
  \bottomrule
  \end{tabular}
  \vspace{-2mm}
  \caption{Experimental results on TKG reasoning (in percentage) under the online setting.}
  \vspace{-2mm}
  \label{table:online_results}
  \end{table*}

    \begin{table}
      \small
      \centering
      \begin{tabular}{lrrrr}
      \toprule
      metrics &MRR  &H@1  &H@3  &H@10\\
      \midrule
      CEN      &42.20 &32.08  &47.46  &61.31\\
      CEN(-CL) &41.50 &31.53  &46.50  &60.81\\
      CEN(-LA) &41.52 &31.49  &46.74  &60.65\\
      \bottomrule
      \end{tabular}
      \vspace{-2mm}
      \caption{Ablation Study of CEN on ICEWS14.}
      \vspace{-2mm}
      \label{table:Ablation}
      \end{table}
\section{Experiments}

{\bf Experimental Setup.} We adopt three widely-used
datasets, ICEWS14~\cite{li2021temporal}, ICEWS18~\cite{jin2020recurrent}, and
WIKI~\cite{leblay2018deriving} to evaluate CEN. Dataset statistics are
demonstrated in Table~\ref{table:datasets}. Due to the space limitation, the
CEN model is only compared with the latest models of TKG reasoning:
CyGNet~\cite{zhu2021learning}, RE-NET~\cite{jin2020recurrent},
xERTE~\cite{han2020explainable}, TG-Tucker~\cite{han2021temporal},
TG-DistMult~\cite{han2021temporal}, TiTer~\cite{sun2021timetraveler} and
RE-GCN~\cite{li2021temporal}. In the experiments, we adopt MRR (Mean Reciprocal
Rank) and Hits@\{1,3,10\} as the metrics for TKG reasoning. We averaged the
metrics over five runs. Note that, following ~\citet{han2020graph}, we adopt an
improved filtered setting where the timestamps of facts are considered,
called time-aware ﬁltered setting. Take a typical query $(s, r, ?, t_{1})$ with
answer $o_{1}$ in the test set for example, and assume there is another two
facts $(s, r, o_{2}, t_{2})$ and $(s, r, o_{3}, t_{1})$. Under this time-aware
filtered setting, only $o_{3}$ will be considered as a correct answer and thus
removed from the ranking list of candidate answers.


{\bf Implementation Details.} In the experiments, the optimal minimum lengths of
evolutional patterns $\hat{k}$ for ICEWS14, ICEWS18, WIKI are 3, 3, 2,
respectively. The maximum length $K$ for all datasets is set to 10. For all
datasets, the kernel width $M$ is set to 3, and $C$ is set to 50. For each fact
$(s, r, o, t)$ in the test set, we evaluate CEN on two queries $(s, r, ?, t)$
and $(?, r, o, t)$. The dimension $d$ of relation representations and entity
representations is set to 200 on all datasets. Adam~\cite{kingma2014adam} is
adopted for parameter learning with the learning rate of 0.001 on all datasets.
The number of RGCN layers is set to 2 and the dropout rate for each layer to
0.2. For the online setting, we set the max epochs of the fine-tuning at each
timestamp to 30. For predicting $G_{t}$, $G_{t-2}$ is used as the validation
set. We fine tune the pre-trained CEN from $T1+1$ to $T_3$ and report the
results at the test timestamps ($T_2$ to $T_3$) in
Table~\ref{table:online_results}. The experiments are carried out on Tesla V100.
Codes are avaliable at https://github.com/Lee-zix/CEN.

\subsection{Experimental Results}\label{Results1} {\bf Results under the Offline
Setting.}  The results under the traditional offline setting are presented in
Table~\ref{table:offline_results}. CEN consistently outperforms the baselines on
MRR, Hits@3, and Hits@10 on all datasets, which justifies the effectiveness of
modeling the evolutional patterns of different lengths. On ICEWS datasets, CEN
underperforms TITer on Hits@1 because TITer retrieves the answer through
explicit paths, which usually gets high Hits@1. Whereas, CEN recalls more answer
entities by aggregating the information from multiple evolutional patterns,
which may be the reason for its high performance on Hits@3 and Hits@10. 







{\bf Results under the Online Setting.}\label{Results2} Under the online
setting, the model is updated via historical facts at the testset. Thus, it
cannot be directly compared with the baselines designed for the offline setting.
As shown in Table~\ref{table:online_results}, on ICEWS datasets CEN outperforms
CEN(-TR) (CEN without TR unit), which implies the effectiveness of TR unit to
balance the knowledge of new evolutional patterns and the existing ones. On
WIKI, CEN(-TR) gets better performance. It is because that the time interval
between two adjacent timestamps in WIKI (one year) is much larger than ICEWS
datasets (one day) and contains more time-variable evolutional patterns. TR unit
limits the model to adapt to new knowledge and is not suitable for this dataset.


{\bf Ablation Study.} To investigate the contributions of curriculum learning
strategy and the length-aware CNN, we conduct ablation studies for CEN on the
test set of ICEWS14 under the traditional offline setting, which are shown in
Table~\ref{table:Ablation}. CEN(-CL) denotes CEN without the curriculum learning
strategy. The underperformance of CEN(-CL) demonstrates the effectiveness of the
curriculum learning strategy. CEN(-LA) denotes the model replacing the
length-aware CNN with a traditional CNN. The underperformance of CEN(-LA)
implies the effectiveness of the length-aware CNN.

\section{Conclusions}
In this paper, we proposed Complex Evolutional Network (CEN) for TKG reasoning,
which deals with two challenges in modeling the complex evolutional patterns:
length-diversity and time-variability. For length-diversity, CEN adopts a
length-aware CNN to learn evolutional patterns of different lengths and is
trained under a curriculum learning strategy. For time-variability, we explored
a new online setting, where the model is expected to be updated to new
evolutional patterns emerging over time. Experimental results demonstrate the
superiority of the proposed model under both the offline and the online
settings.

\section*{Acknowledgments}
The work is supported by the National Natural Science Foundation of China under
grants U1911401, 62002341 and 61772501, the GFKJ Innovation Program, Beijing
Academy of Artificial Intelligence under grant BAAI2019ZD0306, and the
Lenovo-CAS Joint Lab Youth Scientist Project.

\bibliography{anthology,custom}
\bibliographystyle{acl_natbib}



\end{document}